\documentclass{IEEEtran4PSCC}
\ifCLASSINFOpdf
   \usepackage[pdftex]{graphicx}
\else
   \usepackage[dvips]{graphicx}
\fi
\usepackage[cmex10]{amsmath}
\usepackage[cmex10]{amsmath}

\usepackage{cite}
\usepackage{amsmath,amssymb,amsfonts}
\usepackage{algorithmic}
\usepackage{graphicx}
\usepackage{textcomp}
\usepackage{xcolor}
\usepackage{multicol, blindtext}
\usepackage{xspace,empheq,fancybox,amsmath,amssymb,graphicx,epstopdf,epsfig,syntonly,times,amsthm} \usepackage{psfrag,color,bm,array,cite}
\usepackage{url,cite,footnote,xspace,syntonly,algorithm,algorithmic}
\usepackage{verbatim,multirow}
\usepackage[T1]{fontenc}
\usepackage{mathtools}
\usepackage{amsmath,amsfonts,amsthm,bm}
\usepackage{graphicx}
\usepackage{mwe}
\usepackage{caption}
\usepackage{stfloats}
\usepackage{amsfonts}
\usepackage{stackengine}
\usepackage{lipsum}
\usepackage{subfig}

\def\BibTeX{{\rm B\kern-.05em{\sc i\kern-.025em b}\kern-.08em
    T\kern-.1667em\lower.7ex\hbox{E}\kern-.125emX}}
    
\newtheorem{remark}{\bfseries Remark}

\newcolumntype{P}[1]{>{\centering\arraybackslash}p{#1}}

\definecolor{color_yuqi2}{RGB}{234, 135, 47}

\allowdisplaybreaks

\makeatletter
\let\old@ps@headings\ps@headings
\let\old@ps@IEEEtitlepagestyle\ps@IEEEtitlepagestyle
\def\psccfooter#1{%
    \def\ps@headings{%
        \old@ps@headings%
        \def\@oddfoot{\strut\hfill#1\hfill\strut}%
        \def\@evenfoot{\strut\hfill#1\hfill\strut}%
    }%
    \def\ps@IEEEtitlepagestyle{%
        \old@ps@IEEEtitlepagestyle%
        \def\@oddfoot{\strut\hfill#1\hfill\strut}%
        \def\@evenfoot{\strut\hfill#1\hfill\strut}%
    }%
    \ps@headings%
}
\makeatother

\definecolor{color_yuqi}{RGB}{129, 35, 108}

\begin{document}
%
\title{Scalable Learning for Optimal Load Shedding \\ Under Power Grid Emergency Operations}

\author{
\IEEEauthorblockN{Yuqi Zhou, Jeehyun Park, and Hao Zhu}
\IEEEauthorblockA{Department of Electrical and Computer Engineering \\
The University of Texas at Austin
}
}


\maketitle

\begin{abstract}
Effective and timely responses to unexpected contingencies are crucial for enhancing the resilience of power grids. Given the fast, complex process of cascading propagation, corrective actions such as optimal load shedding (OLS) are difficult to attain in large-scale networks due to the computation complexity and communication latency issues. This work puts forth an innovative learning-for-OLS approach by constructing the optimal decision rules of load shedding under a variety of potential contingency scenarios through offline neural network (NN) training. Notably, the proposed NN-based OLS decisions are fully \textit{decentralized}, enabling individual load centers to quickly react to the specific contingency using  readily available  \textit{local measurements}. Numerical studies on the IEEE 14-bus system have demonstrated the effectiveness of our scalable OLS design for real-time responses to severe grid emergency events.
\end{abstract}


\thanksto{\protect\rule{0pt}{0mm} 
This work has been supported by NSF Grants 1802319 and 2130706.
}

\section{Introduction}
Fast mitigation of power imbalance and operational limit violations during emergency events is of great importance for enhancing the resilience  of power grids. To prevent potential cascading failures, load shedding is a commonly used emergency response action by adjusting the system operating point. Unlike normal operations, the decision making of load shedding under severe contingencies is timing-critical. Due to the computational complexity and communication latency concerns arising in attaining load shedding solutions, machine learning (ML) methods are uniquely positioned to enable timely emergency grid services thanks to their superior performance in real-time prediction.

To quickly restore power balance, traditional load shedding schemes perform a uniform percentage of reduction for all load centers based on the electric frequency deviation \cite{concordia1995load}; see also the NERC standard \cite{NERC}. This proportional reduction is easy to implement, but fails to account for the heterogeneous effects of the contingency scenario across the grid such as severe congestion at certain locations. Recent advances in optimization-based load shedding  schemes (e.g., \cite{coffrin2018relaxations,rhodes2020balancing,lin2016admm}) can effectively mitigate the potential risks of cascading failures through strategically targeting locational congestion or stability concerns. Nevertheless, to solve the resultant optimal load shedding (OLS)  problem in real time can be challenged by  the underlying nonlinearity of AC power flow model. Recently, there is a surge of interest in adopting ML methods for power system decision making under normal conditions, particularly for the AC  optimal power flow (AC-OPF) problem; see e.g., \cite{guha2019machine,baker2019learning,deka2019learning}. Similarly, to  accelerate the AC-OLS solution, \cite{kim2019graph} has proposed to learn the percentage ratio of load shedding from the system-wide contingency information.
{In addition, a safe reinforcement learning approach has been developed in \cite{vu2020safe} to predict the dynamic load shedding policy from the overall system state, thus requiring the grid-wide information.}
Although a centralized learning framework is suitable for normal grid conditions, it fails to promptly react to the contingency situation due to the associated communication and response times. 
Therefore, existing ML-based solutions cannot cope with the real-time OLS needs where it is critical to implement timely corrective actions at distributed load centers.

This paper aims to develop a scalable learning-for-OLS framework such that individual load centers can  predict their own optimal decisions in a decentralized fashion. To this end, we first formulate the OLS problem under the AC power flow model, as an extension to the AC-OPF. 
By determining the respective amount of load shedding at each bus, the AC-OLS problem aims to restore the system-wide power balance and mitigate the violations of operational limits. Upon solving this problem under a wide range of loading conditions and contingency scenarios, one can learn the decision rules from input system conditions to target load shedding actions through offline training. In order to attain high accuracy, we adopt the neural network (NN) model to construct nonlinear, expressive mappings from real-time measurements.
One notable feature of our proposed approach is the \textit{decentralized} design of the NN-based decision rules, which have been constructed solely based on  \textit{locally available measurements} at each load center. Thanks to the scalability of the decentralized information sharing, the offline training time to obtain individual decision rules is greatly reduced compared to a full-feedback one. During online emergency operations, these decisions rules further enable each load center to promptly react to the specific contingency from readily available local data.


This paper is organized as follows. 
Section \ref{sec:OLS} formulates the centralized AC-OLS problem. Section \ref{sec:DL} presents the NN training process  for the proposed  decentralized OLS framework.  
Numerical tests on the IEEE 14-bus system are provided in Section \ref{sec:NR} to demonstrate the accurate prediction performance of the proposed solutions to the line outage contingency, and the paper is concluded in Section \ref{sec:con}.

\section{Optimal Load Shedding (OLS) Problem} \label{sec:OLS}

We formulate the optimal load shedding (OLS) problem based on the nonlinear AC power flow. Consider a power grid with $N$ buses and $L$ lines collected in the sets $\cal N$ and $\cal L$, respectively. Let $\bm{Y}= \bm{G} + j \bm{B}$ denote the network admittance matrix, where $\bm{G}$ and $\bm{B}$ are real and imaginary parts, respectively.
For each bus $i$, the complex power output from its connected generation is denoted by $p_{i}^{g}+j q_{i}^{g}$, while load demand by $p_{i}^{d}+jq_{i}^{d}$.  In addition, let $V_{i}$, $\theta_{i}$ denote the bus voltage magnitude and angle, respectively. Thus, the voltage phasor for each bus can be represented in polar form as 
    $ V_{i} \angle \theta_{i}$.
Under the AC power flow, the complex power flow for line $(i,j) \in \cal L$ can be represented as \cite{coffrin2015qc}:
\begin{align} 
    S_{ij} = {Y}_{ij}^{*} V_{i} ^2 - {Y}_{ij}^{*} V_{i} V_{j} \angle(\theta_i-\theta_j)
\end{align}
where $(\cdot)^*$ denotes the conjugate of a complex number.


The AC-OLS problem is formulated similarly to AC optimal power flow (AC-OPF). The latter problem determines the optimal set-points under normal operations that account for system constraints on generation output, voltage, and line flows. Under emergency operations as a result of large-scale contingencies, the AC-OPF problem could become infeasible due to insufficient resources or transfer capability. Thus, corrective actions such as load shedding are typically carried out to maintain power balance and satisfy the system operation limits. With known load demand $p_{i}^{d}$ and $q_{i}^{d}$ per bus $i$, the OLS problem aims to determine  the amount of load shedding denoted by $p_{i}^{s}$ and $q^s_i$, as given by
\begin{subequations} \label{eq:OLS}
\begin{align}
\min \; \, \, & \sum_{i=1}^{N} \, c_{i}^{g}(p_{i}^{g}) + c_{i}^{s}(p_{i}^{s}) \label{eq:OLS_a}\\
\textrm{s.t.} \; \, \, &  p_{i}^{g} \in \mathbb{R}, q_{i}^{g} \in \mathbb{R}, V_{i} \in \mathbb{R}^{+}, \theta_{i} \in \mathbb{R}, \, \forall i \in \cal N \label{eq:OLS_b}\\ 
  & p_{i}^{s} \in \mathbb{R}, q_{i}^{s} \in \mathbb{R}, \, \forall i \in \cal N   \label{eq:OLS_c}\\
  & {\underline{p}_{i}^{g}} \leq p_{i}^{g} \leq {\overline{p}_{i}^{g}}, \: \: {\underline{q}_{i}^{g}} \leq q_{i}^{g} \leq {\overline{q}_{i}^{g}}   \label{eq:OLS_d}\\
  & {\underline{V}_{i}} \leq V_{i} \leq {\overline{V}_{i}}, \: \: {\underline{\theta}_{i}} \leq \theta_{i} \leq {\overline{\theta}_{i}} \label{eq:OLS_e}\\
  & 0 \leq p_{i}^{s} \leq p_{i}^{d}, \: \: 0 \leq q_{i}^{s} \leq q_{i}^{d} \label{eq:OLS_f}\\
  & |S_{ij}| \leq \overline{S}_{ij}  \label{eq:OLS_g}\\
  & \theta_{ij} = \theta_{i} - \theta_{j} \label{eq:OLS_h}\\
  & p_{i}^{g} - p_{i}^{d} + p_{i}^{s} = \textstyle \sum_{j=1}^{N}  V_{i} V_{j} \left(G_{ij}\cos \theta_{ij}  + B_{ij}\sin \theta_{ij}\right)\label{eq:OLS_i} \\
  & q_{i}^{g} - q_{i}^{d} + q_{i}^{s} =  \textstyle \sum_{j=1}^{N} V_{i}V_{j} \left(G_{ij}\sin \theta_{ij}  - B_{ij}\cos \theta_{ij}\right). \label{eq:OLS_j}
\end{align}
\end{subequations}
The objective function in \eqref{eq:OLS_a} consists of the generation cost $c_{i}^{g}(\cdot)$ and load reduction cost $c_{i}^{s}(\cdot)$, which are typically (piece-wise) linear or convex quadratic functions. 
One can also extend the objective to incorporate the cost of reactive power reduction for each load. In general, the
two cost functions $c_{i}^{g}$ and $c_{i}^{s}$ are designed such that for every bus $i \in \cal N$ we have:
\begin{align} \label{eq:3}
  \left|c_{i}^{g}(p_{i}^{g})\right| \ll \left|c_{i}^{s}(p_{i}^{s})\right|.
\end{align}
This way, the OLS solutions will prefer to fully utilize the generation resources before evoking load shedding. 
The load shedding cost can vary from load centers by prioritizing  critical loads with much higher costs than the others.  
Moreover, for the decision variables given in \eqref{eq:OLS_b} - \eqref{eq:OLS_c}, constraints \eqref{eq:OLS_d} - \eqref{eq:OLS_f} enforce the upper/lower bounds for each of them based upon the resource budget or operational limits. The upper bounds for load shedding in \eqref{eq:OLS_f} can be lower than the total demand in case part of the load center constitutes as non-dispatchable critical loads. Furthermore, the constraint \eqref{eq:OLS_g} enforces the line thermal limit for apparent power flow, while other line limits (e.g, line current, real power flow) can be similarly posed. Finally, the equality constraints \eqref{eq:OLS_h} - \eqref{eq:OLS_j} correspond to AC power flow equations. 

The AC-OLS problem differs from the AC-OPF one mainly in the level of flexibility that each load center can provide. During the contingency conditions, corrective actions such as reducing load demand or reconnecting transmission lines are imperative  in order to restore the power balance and increase the power transfer capability. It is worth pointing out that even though the OLS problem \eqref{eq:OLS} is formulated to consider load shedding only, it can be generalized to include other corrective actions such as topology optimization and full de-energization of system components; see e.g., \cite{coffrin2018relaxations,rhodes2020balancing}.

\begin{remark}[Solving AC-OLS]
As a nonlinear program (NLP), the AC-OLS problem \eqref{eq:OLS} is nonconvex and  generally NP-hard, similar to the AC-OPF  \cite{castillo2013survey}. Various convex relaxation methods can be adopted to  tackle the nonconvexity therein for OPF \cite{low2014convex,kocuk2016strong} and similarly for OLS \cite{coffrin2018relaxations}. In general, open-source packages such as MATPOWER \cite{zimmerman2010matpower} and JuMP \cite{coffrin2018powermodels} are available to efficiently solve the OLS problem. 
\end{remark}

\section{Learning the Scalable OLS Strategy} \label{sec:DL}

\begin{figure}[t!]
\centering
\includegraphics[trim=3cm 2.6cm 4cm 3cm,clip=true,width=0.5\textwidth]{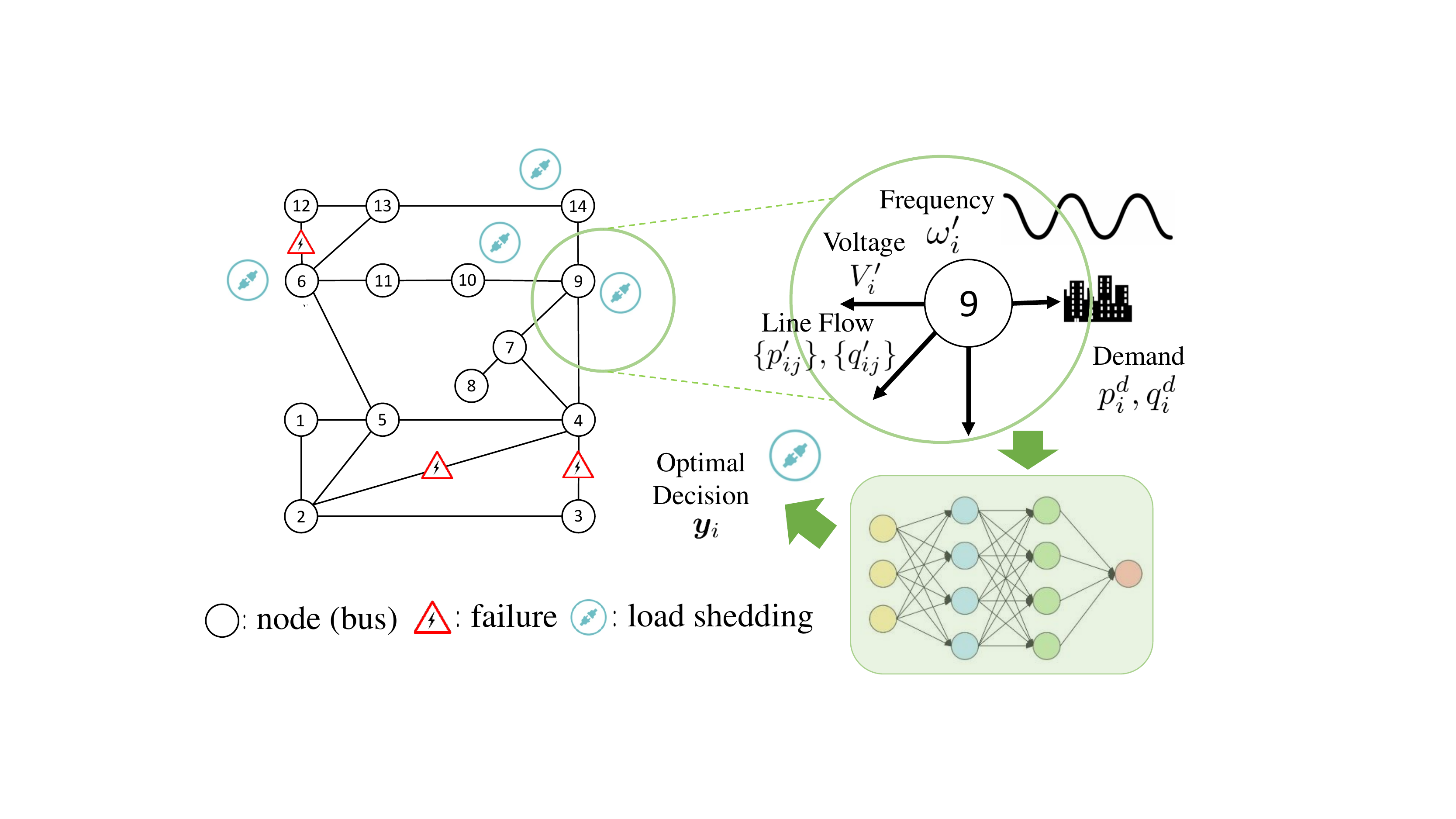}
\caption{Illustration of scalable load shedding design on the IEEE 14-bus system.}
\vspace*{-1mm}
\label{fig:loadshed}
\end{figure}

{Although the centralized OLS problem can be solved by various optimization solvers, its implementation requires high rate of communications for the control center to acquire the system-wide information and dispatch the emergency actions. Due to communication latency and quality issues, this centralized framework could affect the timeliness and effectiveness of corrective action responses at individual load centers, both critical for grid emergency operations.} To enable fast and powerful load shedding actions during emergency events, we propose to develop a scalable OLS strategy by predicting the OLS decision using \textit{locally available}  measurements in real time. 
The key idea of the proposed framework is illustrated in Fig.~\ref{fig:loadshed}, where each load center such as the one in bus 9 can directly form its own OLS decisions using the voltage phasor and power data collected by local meters. Under the physics-based power flow coupling, each contingency scenario leads to various level of changes in the available measurements at every load center. Thus, the latter can be used to infer the OLS solution for the specific contingency even without centralized information exchange.   

\begin{figure}[t!]
\centering
\includegraphics[trim=2.5cm 3cm 2.5cm 3cm,clip=true,width=0.5\textwidth]{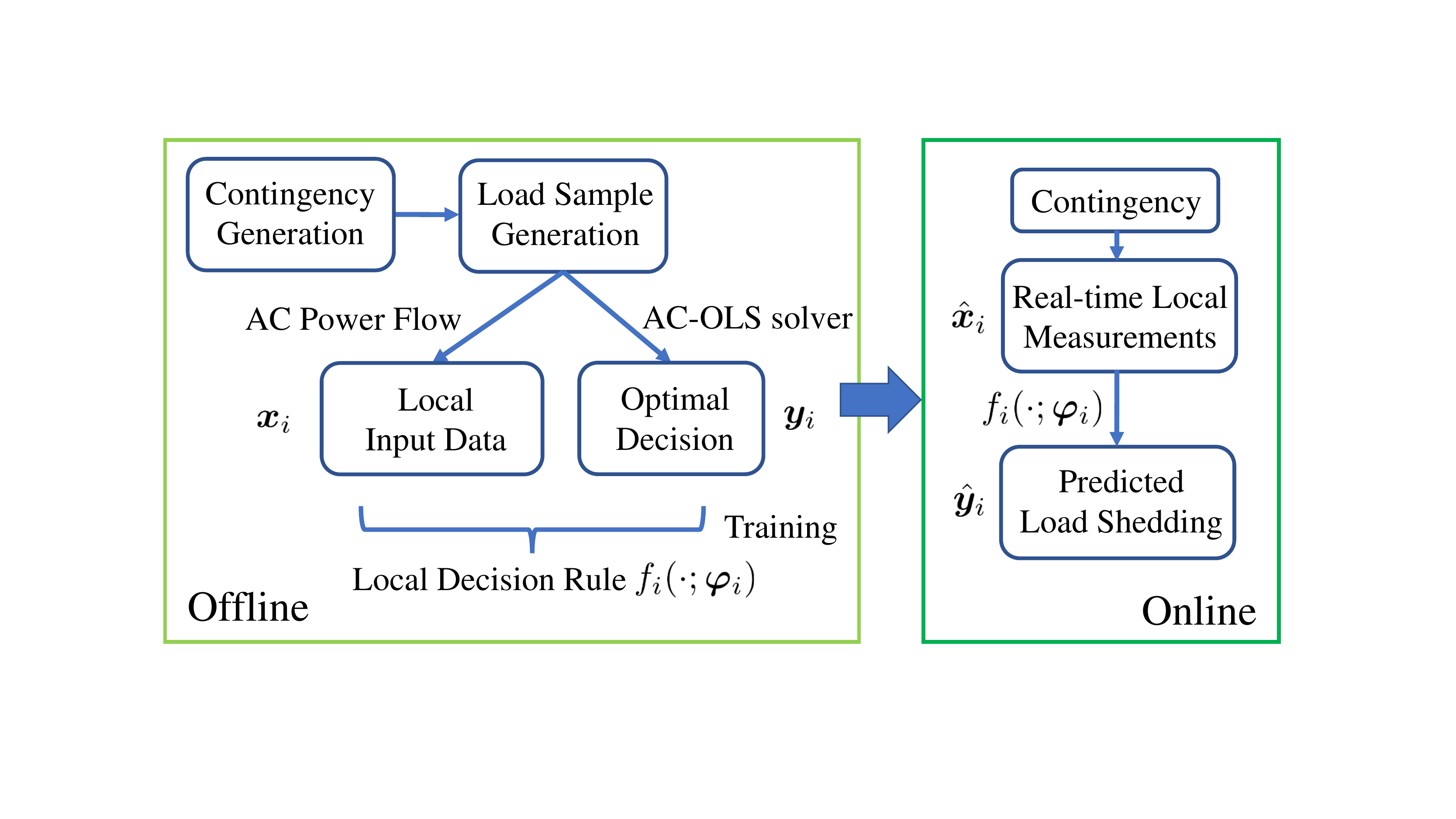}
\caption{The overall architecture for the proposed scalable load shedding design with extensive offline studies and training to accelerate the online decision making.}
\label{fig:framework}
\vspace*{-1mm}
\end{figure}

Of course, the question becomes how to obtain such decentralized decision rules without using system-wide information. To this end, we utilize the basic feedforward neural network to obtain the decision rule $f_i(\bm{x}_{i}; \bm{\varphi}_i)\rightarrow \bm{y}_{i}$ for each load center $i$, that maps from local measurements $\bm{x}_i$ to its optimal decisions $\bm{y}_i$. 
Note that $\bm{\phi}_i$ denotes the NN parameters that will be specified later and  learned during the training process. As the goal is to establish the decision rule $f_i(\cdot; \bm{\varphi}_i)$ for any possible system operating condition or contingency scenario, the offline training process builds upon generating a high number of instances, each representing a specific loading condition and contingency scenario.    
As illustrated in Fig.~\ref{fig:framework}, for each instance, the corresponding input feature $\bm{x}_i$ and target decision $\bm{y}_i$ can be respectively computed by solving the power flow and OLS problem \eqref{eq:OLS}. All these samples of $\{\bm{x}_i,\bm{y}_i\}$ will be used to train the decision rule $f_i$ by learning its parameters $\bm{\varphi}_i$. 
{When using $f_i$ for online implementation, each control center can immediately use
real-time local data $\hat{\bm{x}}_{i}$ to quickly obtain the decision as $\hat{\bm{y}}_{i}=f_i(\hat{\bm{x}}_{i}; \varphi_i)$. 
}
This constitutes the overall architecture of the proposed scalable OLS design, which leverages extensive offline computation and learning to empower the online decision-making process.

For accurate OLS prediction, the local input feature $\bm{x}_i$ should include all possible real-time measurements, such as 
\begin{align} \label{eq:5}
  \bm{x}_{i} = \left[p_{i}^{d}, q_{i}^{d}, V_{i}', \{p_{ij}'\}, \{q_{ij}'\}, \omega_{i}'  \right],
\end{align}
which represents local real/reactive power demand information, post-contingency voltage magnitude, all incident line flows, as well as the electric frequency. Note that symbols with $'$ here correspond to post-contingency values and differ from those in \eqref{eq:OLS}. Most of these values can be computed by open-source solvers such as MATPOWER through steady-state power flow simulations, except for the electric frequency $\omega_i'$. As the latter is a very informative indicator of the overall power imbalance, one can approximate it using the difference between pre- and post-contingency steady-state power  generation  \cite[Ch.~12]{glover2012power}.
In future, we plan to utilize dynamic power flow simulations such as the COSMIC tool \cite{song2015dynamic} to improve the post-contingency modeling.  
As for the target decisions, we are primarily interested in predicting the load reduction solutions  from \eqref{eq:OLS}, as given by:
\begin{align} \label{eq:6}
  \bm{y}_{i} = \left[p_{i}^{s}, q_{i}^{s}   \right].
\end{align}
General corrective decisions can be predicted as well by extending AC-OLS problem  as mentioned earlier.  

The proposed decentralized OLS design builds upon the strong correlation between load shedding decisions and local post-contingency data, as observed in numerical tests later on.  As the measurements in $\bm{x}_{i}$ can effectively reveal the effects of contingency at bus $i$, they are highly indicative of the corresponding optimal decision. For example, both $V_{i}'$ and $\omega_{i}'$ are great stress indicator on the system loading conditions. Similarly, the line flows $\{p_{ij}', q_{ij}'\}$ indicate the change of power flow patterns due to contingency.

To obtain the OLS decision rule $f_i(\cdot; \bm{\varphi}_i)$ for load center $i$, the NN model consists of multiple fully-connected hidden layers between input $\bm x_i$ and output $\bm y_i$. With the first layer  $\bm {z}_i ^0 = \bm x_i$ incorporating the input feature, each layer $k$ can be represented as:
\begin{align} 
\bm{z}^{k}_{i} = \sigma \left(\bm{W}^{k} \bm{z}^{k-1}_{i} + \bm{b}^{k} \right),~~\forall k = 1,\ldots, K
\end{align}
where the final layer $\bm{z}^{K}_{i} \rightarrow \bm y_i$ predicts the output target. Thus, the NN parameters in $\bm{\varphi}_i$ include the weight matrices $\{\bm{W}^{k}\}$ and bias vectors $\{\bm{b}^{k}\}$ for the linear transformation per layer $k$. Each layer also uses a nonlinear activation function  $\sigma (\cdot)$ to attain high-dimensional, expressive functional mapping that goes beyond linearity. Common choices of the activation function include sigmoid and ReLU. 
{To determine $\bm{\varphi}_i$, we use the mean squared error (MSE) metric as the loss function to minimize, solved by popular NN training algorithms such as stochastic gradient descent.
}




\begin{remark}[Safety of decentralized OLS]
As a corrective action, the OLS decisions need to be effective and safe during online implementation. Such considerations can be further incorporated into offline training. For example, a weighted MSE metric could discourage larger prediction error  for higher amount of load shedding amount. In addition, we can generalize it to a risk-aware learning framework using the conditional value-at-risk (CVaR) measure to reduce the worst-case prediction error; see e.g., \cite{lin2021risk}.
\end{remark}

\begin{figure*}[hbt!]
\centering 
   \subfloat[]{
      \includegraphics[trim=0cm 0cm 0cm 0cm,clip=true, width=0.32\textwidth]{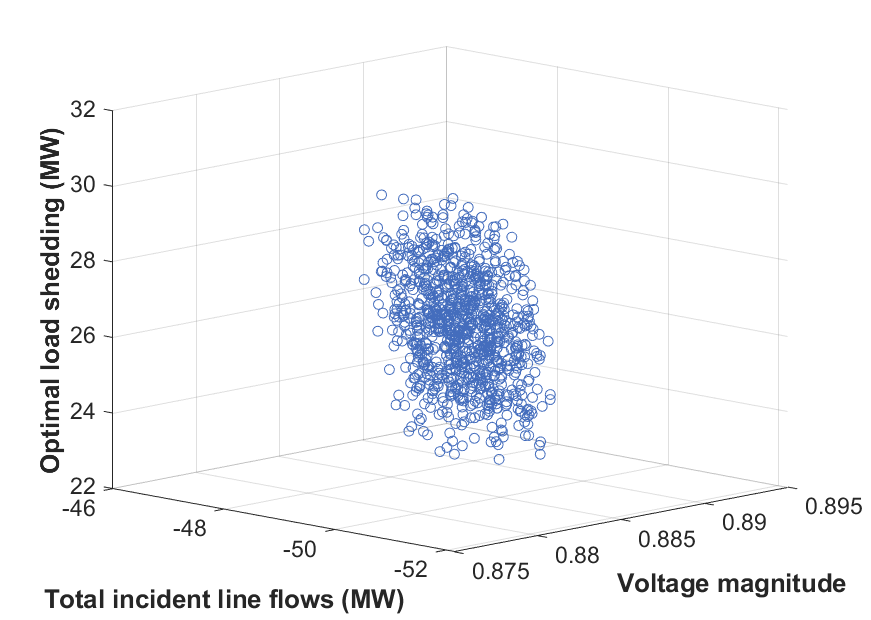}}  
\hspace{\fill}
   \subfloat[]{
      \includegraphics[trim=0cm 0cm 0cm 0cm,clip=true, width=0.32\textwidth]{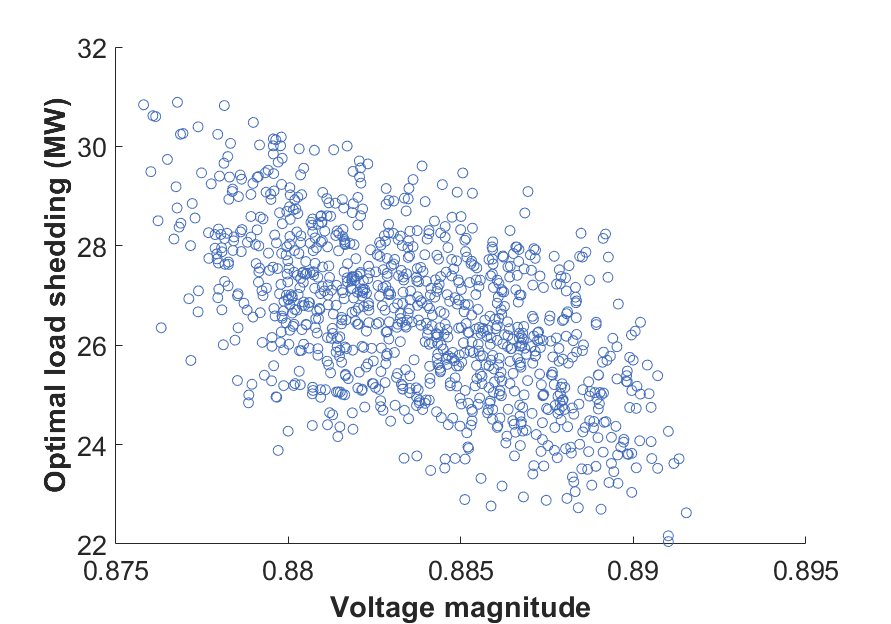}} 
\hspace{\fill}
   \subfloat[]{
      \includegraphics[trim=0cm 0cm 0cm 0cm,clip=true, width=0.32\textwidth]{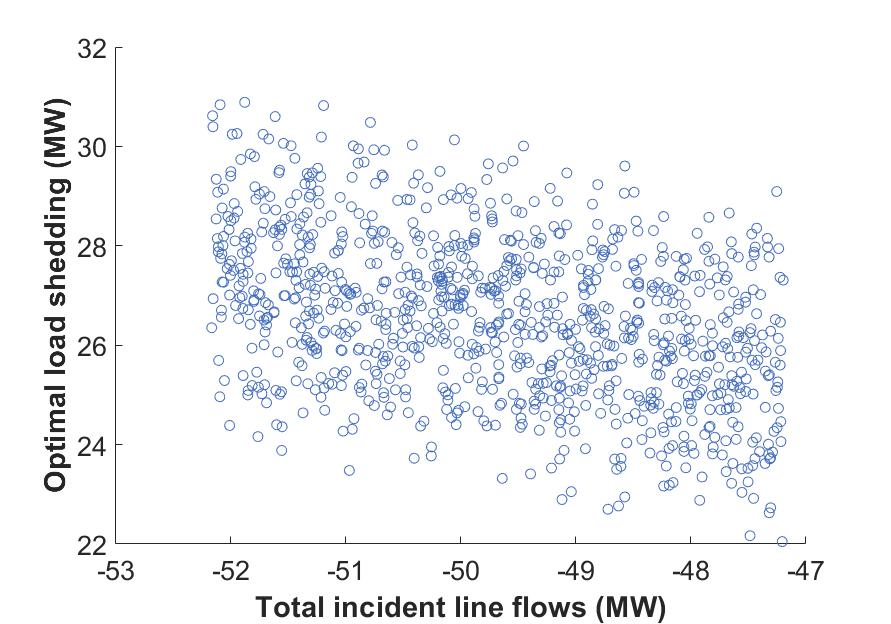}} 
\caption{The optimal load shedding decisions at bus 14 for double-line outages versus (a) both voltage magnitude $V_{i}'$ and total incident line flows $\sum_{i} p_{ij}'$; (b) voltage magnitude $V_{i}'$ only; and (c) total incident line flows $\sum_{i} p_{ij}'$ only.}
\label{fig:scatter}
\vspace*{-5mm}
\end{figure*}

\section{Numerical Validations} \label{sec:NR}
This section presents numerical test results of the proposed decentralized OLS approach on the IEEE 14-bus system. The AC-OLS problem has been implemented using MATPOWER and solved by the primal-dual interior point method. Quadratic objective functions of  $c_{i}^{g}(\cdot)$ and $c_{i}^{s}(\cdot)$ have been used for the cost of generation and shedding loads, respectively.
The feedforward NN has been implemented with MATLAB\textsuperscript{\textregistered} deep learning toolbox using the Bayesian regularization algorithm.
The simulations are performed on a regular laptop with Intel\textsuperscript{\textregistered} CPU @ 2.60 GHz and 16 GB of RAM.

The IEEE 14-bus system is shown as in Fig.~\ref{fig:loadshed}. It consists of 20 transmission lines and 5 conventional generators located at buses 1, 2, 3, 6 and 8.
Given the initial load condition, the load shedding may be required more often at certain buses than the others. To this end, we mainly study the load centers located at buses 6, 9, 10, 11, 13 and 14, respectively.
We consider line outage failures as the initial emergency events, while the proposed method is generalizable to other types of system contingencies as well. 
We have generated all $(N-1)$ contingency scenarios (single), and randomly selected $(N-2)$ and $(N-3)$ contingency scenarios (multiple).
For simplicity, the scenarios that lead to system islanding are excluded here and will be studied in future.
To encourage the occurrence of  system emergency operations, we increase the original system loading to a total of 469 MW, under which the AC-OPF solution is closer to the infeasibility margin and  load shedding is more likely to incur during contingencies. In addition to contingency sampling, we also randomly generate the load demand per bus  to be $[95\%,105\%]$ of its nominal value, to reflect its small variation in minute  time-frame. For each contingency scenario, we generated a total of 1000 samples, a majority of which have experienced the occurrence of AC-OLS due to the stress of emergency conditions.

To demonstrate the correlation between local measurements and OLS decisions, Fig.~\ref{fig:scatter} plots their relations at bus 14 under the outage of both line 2-3 and line 4-9.  The cross-section scatter plots show that the optimal shedding amount increases as the voltage magnitude $V_i'$ or total incident line flow decreases. This is because low voltage indicates system stress,  while reduced incident line flow  implies a change of power flow pattern, both calling for the need of load shedding. This observation supports to use local data to form OLS decisions. Note that for ease of exposition, only the predicted real power reduction amount will be presented. 

\begin{table}[t!] 
\caption{RMSE Values Under Single Line Outage} 
\centering 
\setlength{\tabcolsep}{8pt}
{\renewcommand{\arraystretch}{1.1}
\begin{tabular}{ |c| c | c | c | } 
\hline 
 Bus & Occurrence & Training [MW] & Testing [MW] \\
\hline 
$6$ & 3.6\% & 0.1844 & 0.6091\\  \hline 
$9$ & 5.3\% & 0.0483 & 0.0499\\  \hline 
$10$ & 12.3\% & 0.2559 & 0.5707\\  \hline 
$13$ & 5.3\% & 0.0402 & 0.0433\\\hline 
$14$ & 98.7\% & 0.4085 & 0.4253\\ 
\hline  
\end{tabular} }
\vspace*{-4mm}
\label{tab:1}
\end{table}

\begin{figure}[t!]
\centering
    \centering
    \subfloat[Load center at bus 10]
    {
        \includegraphics[trim=0cm 0cm 0cm 0.8cm,clip=true,totalheight=0.16\textheight]{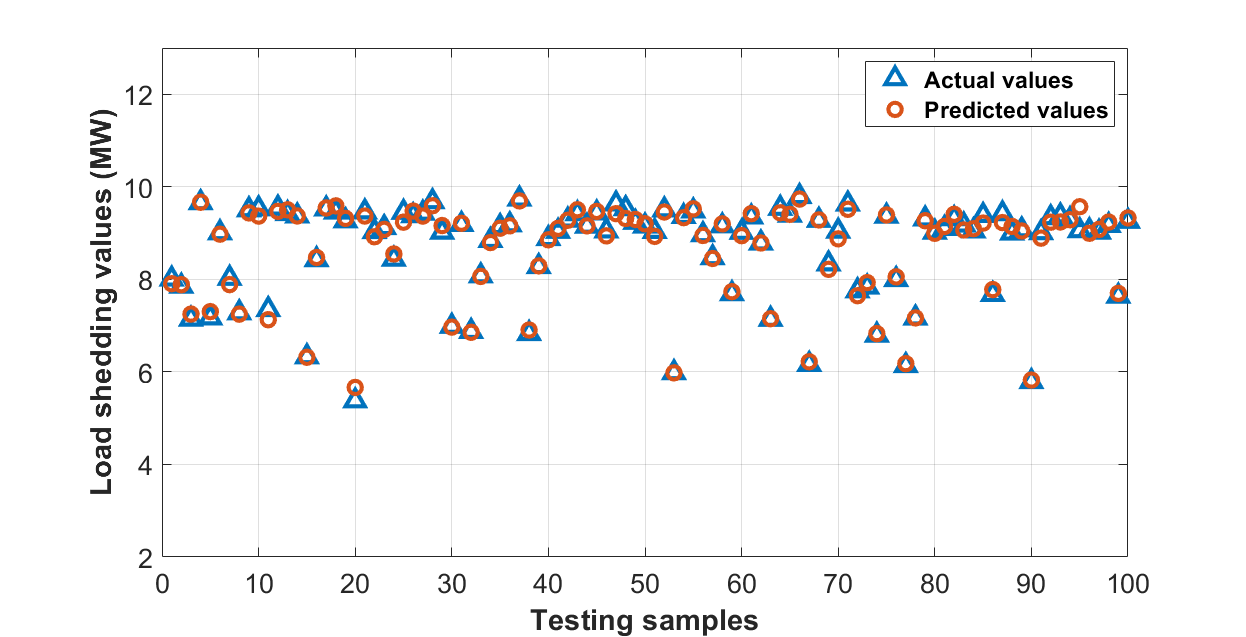}
        \label{fig:single_10_a}
    } \vspace*{-3mm}
    \\
    \centering
    \subfloat[Load center at bus 14]
    {
        \includegraphics[trim=0cm 0cm 0cm 0.8cm,clip=true,totalheight=0.16\textheight]{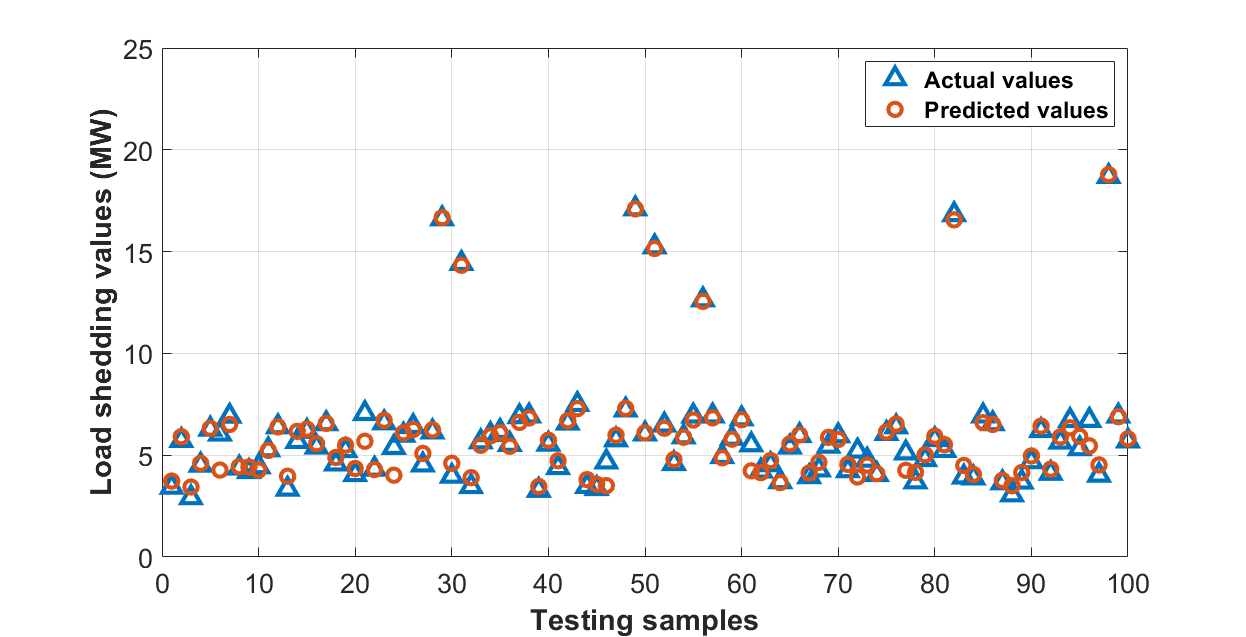}
        \label{fig:single_14_a}
    }
    \caption{Comparison of predicted and actual decision values for randomly selected samples under single line outage testing.}
    \vspace*{-5mm}
    \label{fig:single}
\end{figure}

The training is carried out using a feedforward NN with two hidden layers, each of which has 15 and 12 neurons, respectively. We split the data samples randomly into $80\%$ and $20\%$ for training and testing, separately performed on single and multiple line outages.
On average, the offline training process takes around 16.5 and 38.7 seconds for single and multiple line outages, respectively. As the number of local inputs mainly depends on the incident topology of each load center, the scalability of the training process can be guaranteed under the decentralized design.
\begin{table}[t!] 
\caption{RMSE Values Under Multiple Line Outages} 
\centering 
\setlength{\tabcolsep}{8pt}
{\renewcommand{\arraystretch}{1.1}
\begin{tabular}{ |c| c | c  | c | } 
\hline 
 Bus & Occurrence & Training [MW] & Testing [MW] \\
\hline 
$6$ & 8.0\% & 0.0957 & 0.4716 \\  \hline 
$9$ & 8.0\% & 0.0454 & 0.0490 \\  \hline 
$10$ & 52.7\% & 0.5269 & 0.8282\\  \hline 
$11$ & 2.4\% & 0.0689 & 0.1518\\   \hline 
$13$ & 11.5\% & 0.8140 & 1.4852 \\  \hline 
$14$ & 99.4\% & 0.8517 & 0.9636\\ 
\hline  
\end{tabular} }
\vspace*{-3mm}
\label{tab:2}
\end{table} 
\begin{figure}[t!]
\centering
    \centering
    \subfloat[Load center at bus 10]
    {
        \includegraphics[trim=0cm 0cm 0cm 0.8cm,clip=true,totalheight=0.16\textheight]{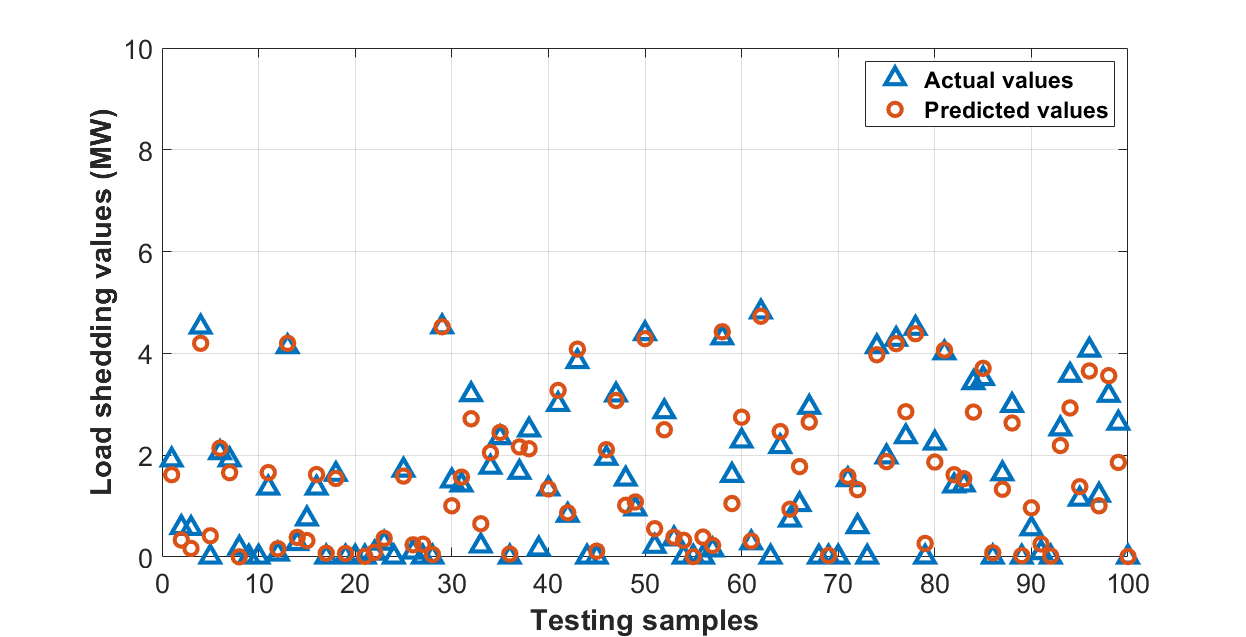}
        \label{fig:multi_10_a}
    } \vspace*{-2.2mm}
    \\
    \centering
    \subfloat[Load center at bus 14]
    {
        \includegraphics[trim=0cm 0cm 0cm 0.8cm,clip=true,totalheight=0.16\textheight]{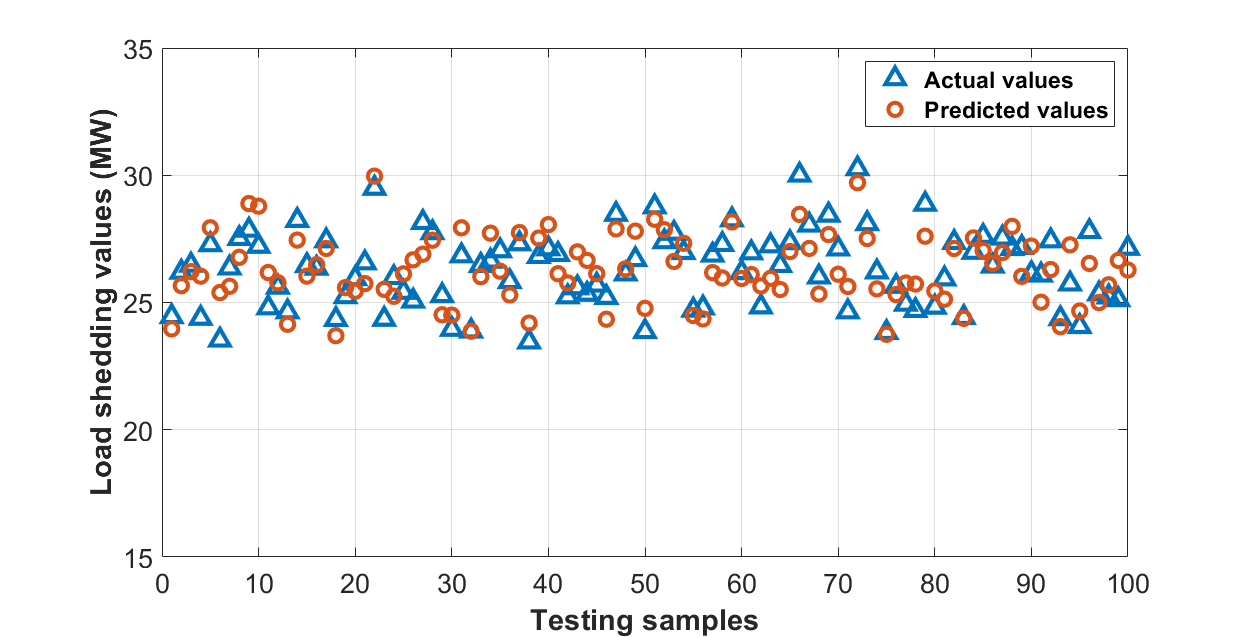}
        \label{fig:multi_14_a}
        
    }
    \caption{Comparison of predicted and actual decision values for randomly selected samples under multiple line outage testing.}
    \label{fig:multiple}
    \vspace*{-5.4mm}
\end{figure}
Table \ref{tab:1} lists the occurrence and prediction error of load shedding at the 5 load buses under single line outage scenarios. Note that bus 11 is not included as load shedding did not happen at this location. As OLS does not occur all the time per bus, we list the root mean square error (RMSE), given by $\sqrt{\frac{1}{S}{\sum_{s=1}^{S} (p_{i}^s - \hat{p}_{i}^s)^{2}}}$, for the samples where OLS has occurred. Fig.~\ref{fig:single} also compares  the predicted and actual load shedding values for selected testing samples at buses 10 and 14, both very likely to need load shedding during  contingencies. 
Note that certain buses such as bus 6 have experienced noticeably higher testing error than  the training one, and we will address this issue through regularization in future. 
Overall, the local predictions well match the actual values of AC-OLS solutions and attain satisfactory performance.

The performance of the proposed design for multiple line outage scenarios is presented in Table \ref{tab:2} and Fig.~\ref{fig:multiple}. As multiple line outages would increase the stress to the system, the occurrence of load shedding has increased with  bus 11 experiencing load shedding in certain cases. By and large, the prediction performance remains good. However, the accuracy has reduced for certain buses compared to the single line outage results, mainly due to the vast variability of post-contingency conditions. Thus, while the simulation results have confirmed  the validity of the proposed approach, we need to expand the variability of the training samples to enhance the expressiveness of the resultant NN models. 


\section{Conclusions and Future Work}
\label{sec:con}
This paper developed a decentralized framework for performing real-time load shedding in order to prevent cascading propagation under emergency events. By solving the AC-OLS optimization problem for a multitude of contingency conditions, we put forth a learning-for-OLS framework that maps from each load center's local measurements to its own OLS decision. Clearly, this scalable design of OLS decision rules enables load centers to quickly react to the contingency situations without requiring the supervision from control center.   Numerical results demonstrate the validity of the proposed design in terms of predicting the OLS solutions.
For future work, we will extensively investigate the proposed design for different contingency conditions and types of systems, as well as improve the safety using risk-aware learning approaches.



%

\bibliography{bibliography.bib}

\bibliographystyle{IEEEtran}

\end{document}